\documentclass[10pt, a4paper]{article}

\usepackage[final]{lrec2026} % this is the new style
\usepackage[utf8]{inputenc}
\usepackage{graphicx}
\usepackage{booktabs}
\usepackage{tablefootnote}
\usepackage{amsmath}
\usepackage{cleveref}
\usepackage{tcolorbox}
\tcbuselibrary{minted, breakable}
\usepackage{minted} % requires -shell-escape
\usepackage{subcaption}
\usepackage{hwemoji}
\newcommand{\flagGER}{🇩🇪} % Germany
\newcommand{\flagITA}{🇮🇹} % Italy
\newcommand{\flagFRA}{🇫🇷} % France
\newcommand{\flagCHN}{🇨🇳} % China
\newcommand{\flagUSA}{🇺🇸} % USA
\newcommand{\flagESP}{🇪🇸} % Spain

\newcommand{\datasetname}{\textsc{PISA-Bench}}
\title{\datasetname{}: The PISA Index as a Multilingual and Multimodal Metric for the Evaluation of Vision-Language Models}

\name{
  Patrick Haller\textsuperscript{1,*},
  Fabio Barth\textsuperscript{2,*},
  Jonas Golde\textsuperscript{1,*},
  Georg Rehm\textsuperscript{2},
  Alan Akbik\textsuperscript{1}
}

\address{
\textsuperscript{1}Humboldt-Universität zu Berlin \quad
\textsuperscript{2}DFKI \\
\textsuperscript{*}Equal contribution}

\abstract{
Vision-language models (VLMs) have demonstrated remarkable progress in multimodal reasoning. However, existing benchmarks remain limited in terms of high-quality, human-verified examples. Many current datasets rely on synthetically generated content by large language models (LLMs). Furthermore, most datasets are limited to English, as manual quality assurance of translated samples is time-consuming and costly. To fill this gap, we introduce \datasetname{}, a multilingual benchmark derived from English examples of the expert-created PISA tests, a unified framework for the assessment of student competencies in over eighty countries. Each example consists of human-extracted instructions, questions, answer options, and images, enriched with question type categories, and has been translated from English into five additional languages (Spanish, German, Chinese, French, and Italian), resulting in a fully parallel corpus covering six languages. We evaluate state-of-the-art vision-language models on \datasetname{} and find that especially small models ($<$20B parameters) fail to achieve high test scores. We further find substantial performance degradation on non-English splits as well as high error-rates when models are tasked with spatial and geometric reasoning. By releasing the dataset and evaluation framework, we provide a resource for advancing research on multilingual multimodal reasoning.
 \\ \newline  }

\begin{document}

\maketitleabstract

\section{Introduction}

Large language models have recently made remarkable progress, demonstrating human-like abilities in tasks such as commonsense question answering \citep{sun2024surveyreasoningfoundationmodels,chen2024llmbasedmultihopquestionanswering,toroghi-etal-2024-right} or mathematical reasoning \citep{wang2025surveylargelanguagemodels,parashar2025inferencetimecomputationsllmreasoning,dong2025scalablellmmathreasoning}. Despite these advances, significant performance differences remain across languages, even in language-agnostic domains such as mathematics.\citep{wu2025bitterlessonlearned2000,Xu_2025}. Further, there are similar disparities across modalities, e.g., where models perform better on visual reasoning tasks when textual information such as image captions or OCR-extracted content is used alongside the image \citep{lu2024mathvista,yue2024mmmumassivemultidisciplinemultimodal}. Developing models capable of reasoning over both images and text in multiple languages is essential to mitigate the current dominance of English-centric systems \citep{zhu2024multilinguallargelanguagemodels}. Such models should be able to, for instance, identify and combine visual and textual information in any language to solve complex reasoning tasks.

However, constructing such datasets is costly, as it requires careful curation of real-world examples that effectively test a model’s ability to reason across text and images. As a result, we currently observe three key limitations in existing benchmarks: (1) many rely on synthetically generated content by LLMs rather than high-quality, human-authored tasks which limits the diversity of the data; (2) multilingual benchmarks often introduce cultural or linguistic biases, limiting the evaluation of a model's true reasoning capability; and (3) most existing datasets are predominantly in English and focus on narrow forms of reasoning, neglecting broader skills such as spatial, geometric, or graph reasoning relevant to education.

To address these issues, we introduce \datasetname{}, a benchmark derived from examples of the PISA tests, an international assessment of student competencies. The PISA test is a large-scale international study conducted by the OECD that evaluates the knowledge and skills of 15-year-old students in reading, mathematics, and science to assess how effectively education systems prepare them for real-world challenges \citep{oecd_pisa}. 

\begin{figure*}[ht!]
\centering
\includegraphics[width=\textwidth]{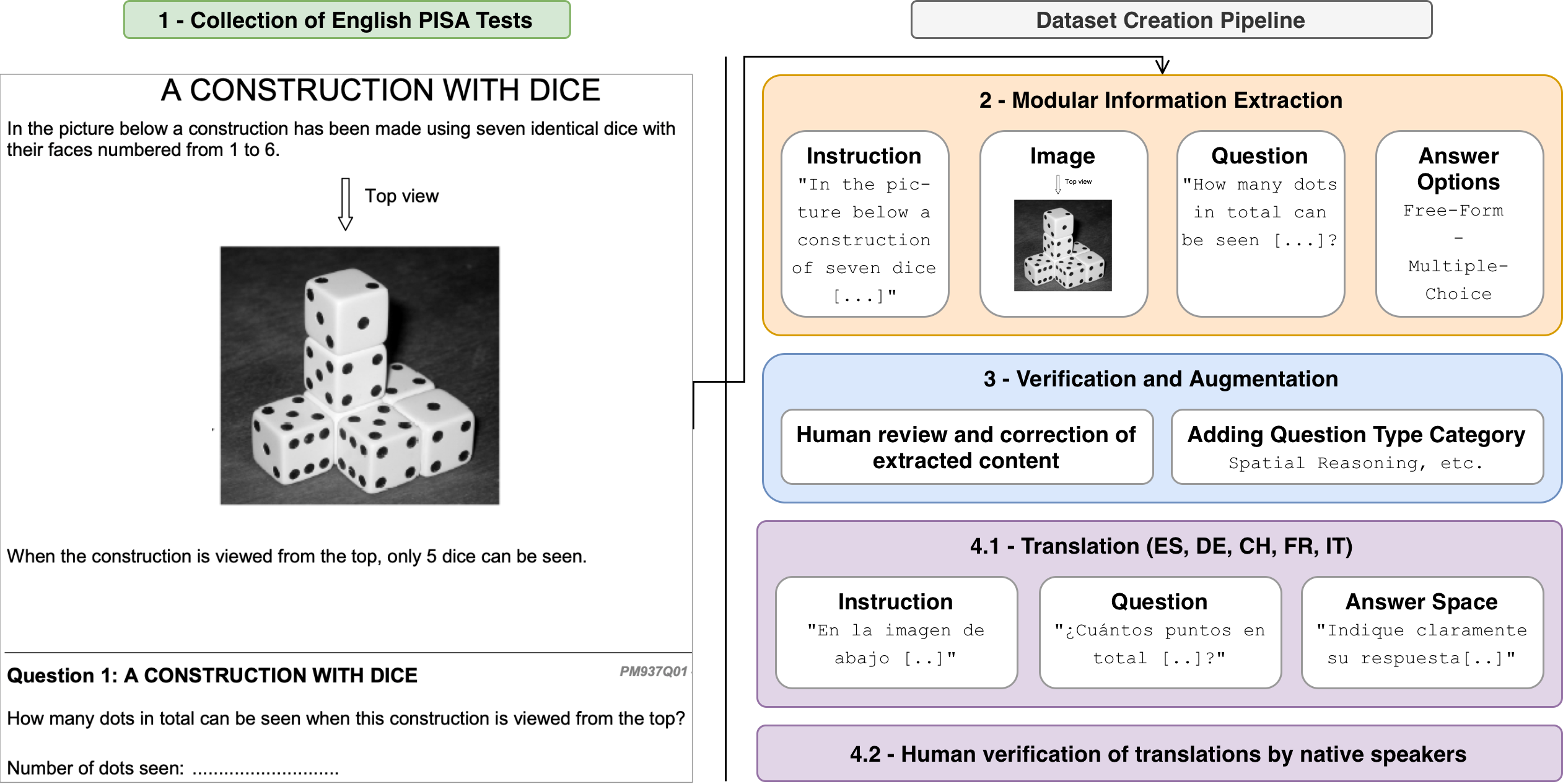}
\caption{
Overview of the dataset construction pipeline. We (1) collect tasks from the original OECD PISA tests, (2) decompose them into modular components (instruction, image, question, and answer options), (3) verify, augment, and, if necessary, correct the extracted content, and (4) translate them into five target languages (ES, DE, CH, FR, IT) and verify translations through native speakers.
}
\label{fig:pisa_pipeline}
\end{figure*}

Our source dataset consists of 122 high-quality test questions. We extract instructions, questions, answer options, and images from the original test documents such that the resulting dataset can be used with a wide range of current state-of-the-art vision-language models. We further enrich each example with metadata, such as question type categories, to identify error categories of the models. Moreover, we generate parallel translations of the English source dataset into five additional languages (Spanish, German, Chinese, French, Italian) to enable multilingual evaluation.

Our main findings show that smaller models with fewer than 20 billion parameters fail to achieve even moderate test scores across all languages (< 55\% on average) while larger and proprietary models achieve moderate accuracies up to 67.8\%. Additionally, for 10 out of the 12 multilingual models tested, we observe average performance drops ranging from -1.4\% to -8.4\% across the translated languages compared to English. Further, we observe that spatial and geometric reasoning is particularly difficult with error rates ranging between 50 and 79\% across languages.

We summarize our contributions as follows:
\begin{enumerate}
    \item We create and release \datasetname{}, a multilingual, parallel benchmark of 122 high-quality examples covering six languages, derived from the PISA tests, including question type categorization,
    \item We conduct extensive evaluation of state-of-the-art vision-language models, revealing significant discrepancies across languages and question types,
    \item We open-source our evaluation framework, enabling future research to easily evaluate their language models in multilingual multimodal reasoning using \datasetname{}.
\end{enumerate}

\section{Dataset Construction}
\label{sec:dataset-construction}

We construct \datasetname{} in a four-stage pipeline. In the first section, we describe the corpus collection using English PISA tests, followed by the modular information extraction. Next, we conduct a quality assurance step in which human annotators verify and, if necessary, correct extracted information, and filtering out those not meeting our quality criteria. At last, we create LLM-based translations into the five target languages which we will verify using human native speakers. We show this process in~\Cref{fig:pisa_pipeline}.

\subsection{Stage 1: Collection of PISA Tests}
We derive the initial corpus from the \textit{Programme for International Student Assessment (PISA)} studies published by the \textit{Organisation for Economic Co-operation and Development (OECD)}.\footnote{\url{https://www.oecd.org/pisa/}} The PISA studies were established to measure how well international education systems prepare students for adult life. They aim to assess not only what 15-year-olds know, but also how effectively they can use that knowledge to solve problems, think critically, and adapt to real-world situations. These materials contain diverse tasks that test students in various categories, including mathematics, science, and reading comprehension. Specifically, some of the examples are not text-only but also include images or figures, which makes them an excellent source of human-created tasks for multimodal reasoning. Furthermore, as the PISA tests aim to compare education systems across countries, they are designed to avoid cultural and linguistic biases. We depict an example in~\Cref{fig:pisa_pipeline} (left), where the test taker must count the number of dice as seen from above, a task that requires spatial reasoning abilities.

To select our source dataset, we collect publicly available PISA tests from 2012 and earlier from the web. Next, human annotators select test questions based on three main criteria: (1) \textit{completeness}, (2) \textit{clarity}, and (3) \textit{multi-modality}. Completeness stands for whether a single example contains a complete instruction (if necessary), a specific question, and answer options (if available). We discard examples that do not meet our standards of clarity, specifically, whether the instruction is ambiguous, for example, requiring the solution of previous tasks. We filter out all text-only examples, such that only multimodal questions are included. We consider examples from the 2012 and earlier PISA tests, and after filtering, our dataset consists of 122 examples.

\subsection{Stage 2: Modular Information Extraction} \label{sec:modular_info_extraction}
We standardize each example by converting it into a structured and modular format. Our annotators extract the following fields:
\begin{itemize}
    \item \textbf{Instruction:} The contextual description provided to the student that introduces to the overall topic or task.  
    \item \textbf{Image:} The corresponding visual material, such as images or figures.  
    \item \textbf{Question:} The main problem statement and question the student needs to answer.
    \item \textbf{Answer Options:} The expected response type, categorized as free-form generation or multiple-choice.  
\end{itemize}

A single PISA question may consist of multiple subquestions. We treat each subquestion as an independent example and, when necessary, augment it with the relevant task instructions to ensure it is self-contained and does not depend on solutions or information from preceding subtasks.

We show the extracted types exemplarily in~\Cref{fig:pisa_pipeline} (cf. 2 - Modular Information Extraction). We extract instruction, image, question, and answer options, which is, in this case, a free-form answer generation. At last, we will also extract the solution section of the original test for the gold answer. To ensure consistency and self-containment, we use GPT-4o \citep{openai2024gpt4ocard} to (1) generate possibly missing multiple-choice options and (2) rephrase questions so that they can be answered either by selecting a multiple-choice option or by generating a free-form response. We show this prompt used for this step in in the ~\Cref{sec:extraction_prompt_appendix}. 
This allows us to evaluate our benchmark in three settings: (1) log-likelihood-based ranking of answer options, as commonly implemented in the LM Evaluation Harness \citep{eval-harness}; and free-form answer generation evaluated using either (2) string matching or (3) more advanced techniques such as LLM-as-a-judge \citep{zheng2023judgingllmasajudgemtbenchchatbot} or sentence similarity \citep{zhang2020bertscoreevaluatingtextgeneration}.

We further label each question with question type categories to analyze the errors of the model. We also use GPT-4o to classify each example in our benchmark into one of the following categories: spatial and geometric reasoning, quantitative reasoning, graph and pattern analysis, and text and diagram understanding. The example in~\Cref{fig:pisa_pipeline} illustrates a spatial and geometric reasoning task, which requires interpreting the dice construction.

\subsection{Stage 3: Quality Control}

After generating the initial dataset, human annotators manually review all materials to ensure that only fully specified tasks remain. Each sample was checked twice by two independent annotators. They checked the following criteria: 
\begin{itemize}
    \item The question should only be answerable using the image.   
    \item The question should closely resemble the original questions' content and intent.
    \item The question should not contain the answer.
    \item The answer options should be reasonable and not trivial.
    \item The text should be in fluent English without syntactic or grammatical mistakes.
\end{itemize}

If any of the criteria are not fulfilled, the sample is regenerated or modified accordingly to fulfill all criteria and ensure a high-quality base dataset in English. This process yields a corpus of \textit{122} high-quality English examples.

\subsection{Stage 4: Translation into Target Languages} \label{sec:translation_target_languages}
We translate the questions and multiple-choice options into five target languages using GPT-4 to enable multilingual evaluation. We show the translation prompt in~\Cref{sec:translation_prompt_appendix}.

We keep all images in their original English versions to preserve comparability across languages. 
%While translating the visual elements may improve linguistic alignment, we also consider two challenges it would introduce: (1) many images contain measurement units or cultural references that are difficult to adapt consistently, and (2) full metric conversions could lead to semantic drift. Given these downsides, we decide to retain the original images in English and thus maintain a consistent visual context and ensure fair evaluation across all language versions.

\begin{table*}[ht]
\centering
\begin{tabular}{lcccc}
\toprule
\textsc{Language} & \textsc{Error-Free (\%)} & \textsc{Critical Mistakes} & \textsc{Major Mistakes} & \textsc{Minor Mistakes} \\
\midrule
Chinese & 66.37 & 21 & 4 & 13\\%0 & 25 & 13 \\
German  & 64.60 & 19 & 10 & 11\\%0 & 29 & 11 \\
French  & 71.68 & 15 & 8 & 9\\%0 & 23 & 9 \\
Italian & 75.22 & 15 & 3 & 10\\%0 & 25 & 10 \\
Spanish & 76.99 & 10 & 5 & 11\\%0 & 15 & 11 \\
\bottomrule
\end{tabular}
\caption{GEMBA-MQM translation validation results using GPT-4 as evaluator.}
\label{tab:gemba_mqm}
\end{table*}

\section{Translation Validation}
\label{sec:translation-validation}

As described in~\Cref{sec:translation_target_languages}, we translated the English source material into five languages (German, Spanish, French, Italian, and Chinese). We validate the translation quality of \datasetname{} using automatic evaluation and human verification to ensure a reliable multilingual benchmark. For automatic validation, we use two metrics: the WMT23 COMET-KIWI \cite{rei2022cometkiwiistunbabel2022submission} and the GEMBA-MQM \cite{kocmi2023gembamqmdetectingtranslationquality} using GPT-4. For human verification, we work with a native speaker in the respective language who verifies 50 random examples for linguistic accuracy. This quality assurance process aims to ensure that all translations in \datasetname{} are both linguistically accurate and semantically faithful to the original content.

\subsection{Automatic Validation}

In this section, we analyze the WMT23 COMET-KIWI and the GEMBA-MQM metrics using GPT-4. 

\paragraph{WMT23 COMET-KIWI Metric.}~The WMT23 COMET-KIWI metric is commonly used for translation validation using a regression-based multilingual transformer. The metric is calculated using a reference text and a machine-translated input to compute a score between 0 and 1, where higher values indicate greater semantic alignment between the reference and translated text. The pretrained WMT23 COMET-KIWI models support over 90 languages, including the six languages of our benchmark.

\begin{figure}[ht!]
\centering
\includegraphics[scale=0.5]{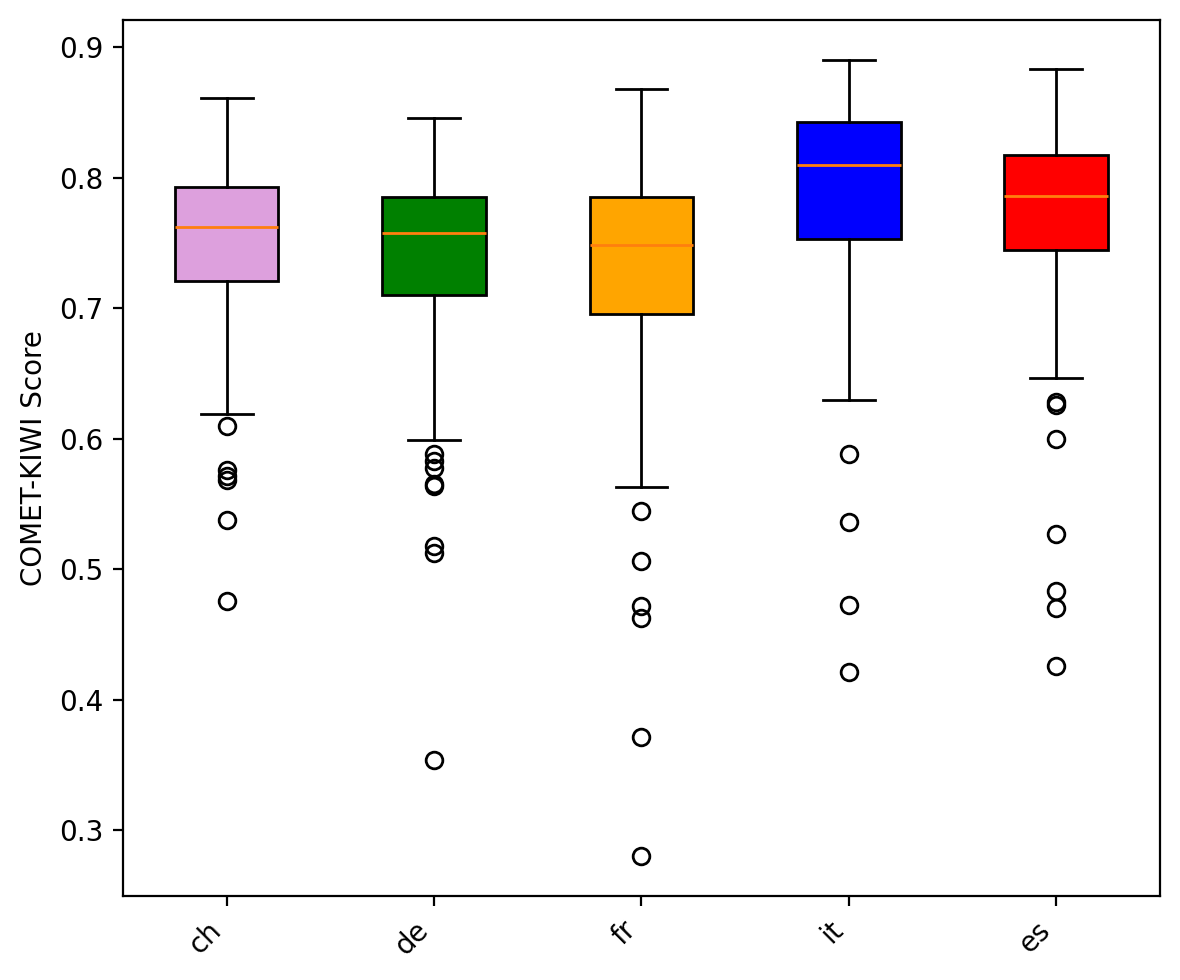}
\caption{Distribution of WMT23 COMET-KIWI scores for each target language.}
\label{fig:comet_boxplot}
\end{figure}

We show the WMT23 COMET-KIWI results in~\Cref{fig:comet_boxplot}. We observe that boxes are consistently above 0.7, indicating good overall translation quality, with Italian and Spanish achieving the highest medians (above 0.8), followed by Chinese, German, and French. Only the box of the French translation goes slightly below 0.7. We further note that each of the translations has some outliers yielding scores below 0.5. Notably, French has data points going down until 0.3, indicating that there may be outliers with low translation quality. Overall, we find that the majority of translations achieve a sufficient level of quality (above 0.7).

\begin{table*}[ht]
\centering
\begin{tabular}{lcccc}
\toprule
\textsc{Language} & \textsc{Error-Free (\%)} & \textsc{Critical Mistakes} & \textsc{Major Mistakes} & \textsc{Minor Mistakes} \\
\midrule
Chinese & 86.00 & 0 & 0 & 6 \\
German  & 86.00 & 0 & 2 & 4 \\
French  & 88.00 & 0 & 0 & 3 \\
Italian & 82.00 & 0 & 0 & 9 \\
Spanish & 76.00 & 0 & 3 & 8 \\
\bottomrule
\end{tabular}
\caption{Human verification results on a random subsample of 50 translated examples.}
\label{tab:manual_verification}
\end{table*}

\paragraph{GEMBA-MQM Metric.}

GEMBA-MQM uses an autoregressive language model, such as GPT, to detect translation quality errors without the need for human reference translations. It classifies the translated text, more specifically the translated spans within the text, into the following three categories: critical issue, major issue, and minor issue. Specifically, we use GPT-4 as the evaluator and adopt the suggested hyperparameters, including the three-shot prompting setup. We reuse the few-shot examples provided by the authors of GEMBA-MQM. 

We show results for GEMBA-MQM evaluation in~\Cref{tab:gemba_mqm}. Specifically, we report the errors per category as well as the error-free rate, defined as the proportion of examples in which no error has been detected. We observe that the results confirm the previous quantitative findings, showing moderate to high error-free rate across all languages (66.3\% - 76.9\%). From a qualitative perspective, we observe a moderate number of critical, major, and minor translation errors, ranging from 10 to 21 critical errors (for Spanish and Chinese, respectively), 3 to 10 major errors (for Italian and German), and 9 to 13 minor errors (for French and Chinese). Italian and Spanish show the highest error-free rates and the lowest absolute errors, while French follows closely. The distribution of critical, major, and minor mistakes suggests that translation quality is consistent, with only a small number of semantic distortions or omissions. However, the authors of GEMBA-MQM point out that the metric should be used with caution in academic comparisons due to its reliance on proprietary, black-box models.

\subsection{Human Validation}

Finally, we work with native speakers who manually review a random subset of 50 translated items (approximately~41\% of the dataset). Each reviewer evaluates the translations according to two criteria:
\begin{itemize}
\item \textit{completeness}, ensuring that no content is omitted, and 
\item \textit{correctness}, ensuring that meaning is preserved without distortion. 
\end{itemize}
The annotator guidelines for verifying the samples are similar to MQM, in that they categorize any translation errors into major and minor categories. We define critical mistakes as cases where sentences contain more than one major error.

We present the results in~\Cref{tab:manual_verification}. They generally show higher error-free rates than those calculated by GEMBA-MQM, indicating the translations made by GPT-4 are of good quality and GEMBA-MQM might be overly critical. We observe that all languages do not contain any critical issues and only a few major issues. However, they do have a small fraction of examples containing minor issues. Spanish has the highest amount of major translation errors, with three examples, which is still in a reasonable range. Finally, we observe that all translations show an error-free rate of at least 76\%.

%\subsection{Verification Procedure and Quality Control}

% zwei phasen quality control auf den englishcen source daten
%We follow a two-step verification procedure.  
%Translations are first produced automatically and then reviewed by native speakers.  
%Annotators inspect each text for completeness and correctness and decide whether to accept, correct, or reject the translation. If issues are identified, we iterate until the annotator confirms the final version.  
%Minor corrections, such as grammatical adjustments, are directly applied to preserve naturalness while maintaining semantic equivalence.

\section{Experimental Setup}
\label{sec:experimental-setup}
In this section, we evaluate several open-weight vision-language models (VLMs) on \datasetname{} to assess the difficulty of our benchmark. We first evaluate on the original English split to establish a baseline performance. We then continue with evaluating all translated versions of our benchmark. At last, we conduct a contamination and error analysis as supporting ablations.

\paragraph{Evaluation Protocol.} 
For each example in our dataset, we provide the model with the instruction, question text, and the associated image. We apply the LLM-as-a-judge protocol with GPT-4 as the evaluator; thus, all models are tasked to generate a free-form textual answer, which is compared to the extracted gold reference.

\paragraph{Models.} 
We consider the following models for our experiments: Qwen2.5-VL (3B -- 72B) \citep{bai2025qwen25vltechnicalreport}, Qwen3-VL \citep{qwen3technicalreport}, Gemma-3 (4B -- 27B) \citep{gemmateam2025gemma3technicalreport}, LLaVA (7B -- 34B) \citep{liu2023visualinstructiontuning}, and Idefics3 (8B) \citep{laurençon2024mattersbuildingvisionlanguagemodels} to cover different architectural concepts, pretraining objectives and model sizes.
We further include GPT-4o and Claude-3.5-Haiku \citep{anthropic2025_claude35haiku} as proprietary models.

\section{Evaluation Results}
\label{sec:evaluation-results}

\paragraph{English Evaluation.} 
\Cref{tab:multilingual_results} summarizes the results on \datasetname{}.  
First, we observe that performance on the English split varies considerably across model sizes, e.g., Qwen2.5-VL (72B) achieves +22.6 accuracy compared to the 3B counterpart. This observation also holds for other model families considered (Gemma3 (27B) vs. Gemma3 (4B): +25.3, LLaVA (34B) vs. LLaVA (7B): +12.9). Qwen2.5-VL (72B) is the only open-source model that is close to proprietary models, only being -1.6 accuracy behind GPT-4o but beating Claude-3.5 Haiku by +6.5. Further, we find that dense models perform better when comparing Qwen3-VL 8B (58.9) and its 30B counterpart with 3B active parameters (57.3). However, we also observe one outlier with Qwen2.5-VL (32B) which does not achieve better performance than its smaller 7B-version.

When comparing model families, we see that Qwen2.5-VL, Qwen3-VL, and Gemma3 achieve significantly better results than LLaVA or Idefics3. Especially when comparing small model scale (up to 8B), we observe that Qwen2.5-VL (3B) and Qwen3-VL (4B) outperform LLaVA (7B) by +16.2 and 17.8 accuracy, respectively, and perform comparatively to Idefics3 (8B). One possible explanation is that models like, e.g., Qwen2.5-VL  and Qwen3-VL have been trained on 4T and 36T tokens, respectively.

\begin{table*}[t]
\centering
\small
\resizebox{\textwidth}{!}{
\begin{tabular}{lrrrrrrr|r}
\toprule
\textsc{Model} & \textsc{en} & \textsc{de} & \textsc{fr} & \textsc{it} & \textsc{es} & \textsc{ch} & \textsc{Avg} & $\Delta_{\text{non-EN}}$ \\
\midrule
\textit{Proprietary Models} \\
GPT-4o & \textbf{71.0±4.1} & \textbf{68.9±4.2} & \textbf{69.7±4.2} & \textbf{65.6±4.3} & \textbf{64.8±4.3} & \textbf{67.2±4.3} & 67.8 & -3.8 \\
Claude-3-5-Haiku & 62.9±4.3 & 56.6±4.5 & 64.8±4.3 & 59.8±4.4 & 61.5±4.4 & 64.8±4.3 & 61.7 & -1.4 \\

\midrule
\textit{Qwen2.5 VL Family} \\
Qwen2.5-VL-3B-Instruct & 46.8±4.5 & 41.0±4.5 & 42.6±4.5 & 43.4±4.5 & 41.0±4.5 & 40.2±4.4 & 42.5 & -5.1 \\
Qwen2.5-VL-7B-Instruct & 52.4±4.5 & 48.4±4.5 & 56.6±4.5 & 54.1±4.5 & 46.7±4.5 & 47.5±4.5 & 50.9 & -1.8 \\

Qwen2.5-VL-32B-Instruct & 51.6±4.5 & 44.3±4.5 & 46.7±4.5 & 45.1±4.5 & 39.3±4.4 & 44.3±4.5 & 45.2 & -7.7 \\
Qwen2.5-VL-72B-Instruct & 69.4±4.1 & 58.2±4.5 & 60.7±4.4 & 64.8±4.3 & 63.1±4.4 & 63.9±4.3 & 63.3 & -7.2 \\
\midrule
\textit{Qwen3 VL Family} \\
Qwen3-VL-4B-Instruct & 48.4±4.5 & 50.0±4.5 & 51.6±4.5 & 49.2±4.5 & 45.1±4.5 & 52.5±4.5 & 49.5 & +1.3 \\
Qwen3-VL-8B-Instruct & 58.9±4.4 & 52.5±4.5 & 57.4±4.5 & 48.4±4.5 & 55.7±4.5 & 55.7±4.5 & 54.8 & -4.9 \\
Qwen3-VL-30B-A3B-Instruct & 57.3±4.4 & 48.4±4.5 & 50.8±4.5 & 50.8±4.5 & 44.3±4.5 & 50.0±4.5 & 50.3 & -8.4 \\

\midrule
\textit{Gemma Family} \\
gemma-3-4b-it & 45.2±4.5 & 35.2±4.3 & 36.9±4.4 & 36.9±4.4 & 36.1±4.3 & 38.5±4.4 & 38.1 & -8.4 \\
gemma-3-12b-it & 58.1±4.4 & 52.5±4.5 & 50.8±4.5 & 51.6±4.5 & 48.4±4.5 & 54.1±4.5 & 52.6 & -6.6 \\
gemma-3-27b-it & 60.5±4.4 & 63.9±4.3 & 61.5±4.4 & 63.9±4.3 & 61.5±4.4 & 54.1±4.5 & 60.9 & +0.5 \\
\midrule
\textit{SmolLM Family} \\
SmolVLM2-256M-Video-Instruct & 33.9±4.3 & 36.9±4.4 & 41.8±4.5 & 41.8±4.5 & 44.3±4.5 & 31.1±4.2 & 38.3 & +5.3 \\
SmolVLM2-500M-Video-Instruct & 29.8±4.1 & 23.0±3.8 & 13.1±3.1 & 17.2±3.4 & 19.7±3.6 & 20.5±3.7 & 20.5 & -11.2 \\
SmolVLM2-2.2B-Instruct & 38.7±4.4 & 23.8±3.9 & 25.4±3.9 & 29.5±4.1 & 34.4±4.3 & 30.3±4.2 & 30.4 & -10.0 \\
\midrule
\textit{LLaVA Family} \\
llava-1.5-7b-hf & 30.6±4.1 & 29.5±4.1 & 32.8±4.3 & 36.1±4.3 & 31.1±4.2 & 29.5±4.1 & 31.6 & +1.2 \\
llava-1.5-13b-hf & 35.5±4.3 & 32.8±4.3 & 27.0±4.0 & 31.1±4.2 & 28.7±4.1 & 33.6±4.3 & 31.5 & -4.8 \\
llava-v1.6-34b-hf & 43.5±4.5 & 36.9±4.4 & 36.9±4.4 & 34.4±4.3 & 38.5±4.4 & 41.8±4.5 & 38.7 & -5.8 \\
\midrule
\textit{Others} \\
Idefics3-8B-Llama3 & 47.6±4.5 & 42.6±4.5 & 36.9±4.4 & 38.5±4.4 & 42.6±4.5 & 36.9±4.4 & 40.9 & -8.1 \\
\bottomrule
\end{tabular}}
\caption{Accuracy (\%) across languages for each model. Best per language in bold. Avg = mean across languages. $\Delta_{\text{non-EN}}$ = Avg(non-EN) - EN in percentage points. CH = Chinese, DE = German, EN = English, ES =
Spanish, FR = French, IT = Italian. Output correctness decided by LLM (chatgpt-4o-mini)}
\label{tab:multilingual_results}
\end{table*}

\begin{table}[t]
\centering
\small
\resizebox{\linewidth}{!}{
\begin{tabular}{lccc}
\toprule
\textsc{Model} & \textsc{mmlu} & \textsc{mmmu} & \textsc{pisa}$_{en}$ \\
\midrule
\textit{Qwen2.5 VL Family} \\
Qwen2.5-VL-3B-Instruct & 66.3 & 46.1 & 46.8 \\
Qwen2.5-VL-7B-Instruct & 68.5 & 52.4 & 52.4 \\
Qwen2.5-VL-32B-Instruct & 75.2 & 58.2 & 51.6 \\
Qwen2.5-VL-72B-Instruct & 83.0 & 70.2 & 69.4 \\

\midrule
\textit{Qwen3 VL Family} \\
Qwen3-VL-4B-Instruct & 64.6 & 54.0 & 48.4 \\
Qwen3-VL-8B-Instruct & 66.6 & 54.8 & 58.9 \\
Qwen3-VL-30B-A3B-Instruct & 66.3 & 59.0 & 57.3 \\

\midrule
\textit{Gemma Family} \\
gemma-3-4b-it & 52.4 & 39.8 & 45.2 \\
gemma-3-12b-it & 72.1 & 48.7 & 58.1 \\
gemma-3-27b-it & 74.6 & 52.0 & 60.5 \\

\midrule
\textit{LLaVA Family} \\
llava-1.5-7b-hf & 25.4 & 33.1 & 30.6 \\
llava-1.5-13b-hf & 25.4 & 35.7 & 35.5 \\
llava-v1.6-34b-hf & 25.4 & 50.2 & 43.5 \\

\midrule
\textit{Others} \\
Idefics3-8B-Llama3 & 23.1 & 46.6 & 47.6 \\
\bottomrule
\end{tabular}}
\caption{Accuracy by benchmark for each model. For \datasetname{}, the English split is shown.}
\label{tab:bench-compare}
\end{table}

\paragraph{Multilingual Evaluation.}
In this section, we discuss evaluation results on the translated splits of our benchmark and show results in~\Cref{tab:multilingual_results}.

First, we observe that most results on the translated version are lower than for the English one. For instance, Qwen2.5-VL (72B) achieves 69.4 accuracy on the English split but only 58.2 on the German split. However, there are also a few outliers, e.g., Gemma3 (27B) achieves 60.5 accuracy on English, but it achieves better performance on German (63.9), French (61.5), Italian (63.9), and Spanish (61.5). Only when evaluating on the Chinese split, we observe lower scores compared to the English split, with 54.1 accuracy. This trend also holds for the LLaVA model family and Idefics3; however, we note that these models are not explicitly trained on multilingual data. The best performing model across languages is GPT-4o again, achieving, e.g., 67.2 accuracy on Chinese or 69.7 on French. The other proprietary model we investigated is the smaller Haiku model of the Claude family, which achieves 61.7 on average, performing slightly worse (-1.6pp.) than the best-performing open-source model, Qwen2.5-VL (72B).

For easier comparison, we included the column $\Delta_{\text{non-EN}}$, which shows the absolute difference between English results and the average across all translated splits. We observe performance decreases of up to -8.4pp. in accuracy on Gemma3 (4B). However, we also observe that the larger version of Gemma3 (27B) achieves better performance on average (+0.5pp.) compared to the English is split, which is the only one together with LLaVA (7B) across all models evaluated.

\begin{figure*}[ht]
  \centering
  \begin{subfigure}[b]{0.48\textwidth}
    \centering
    \includegraphics[width=\textwidth]{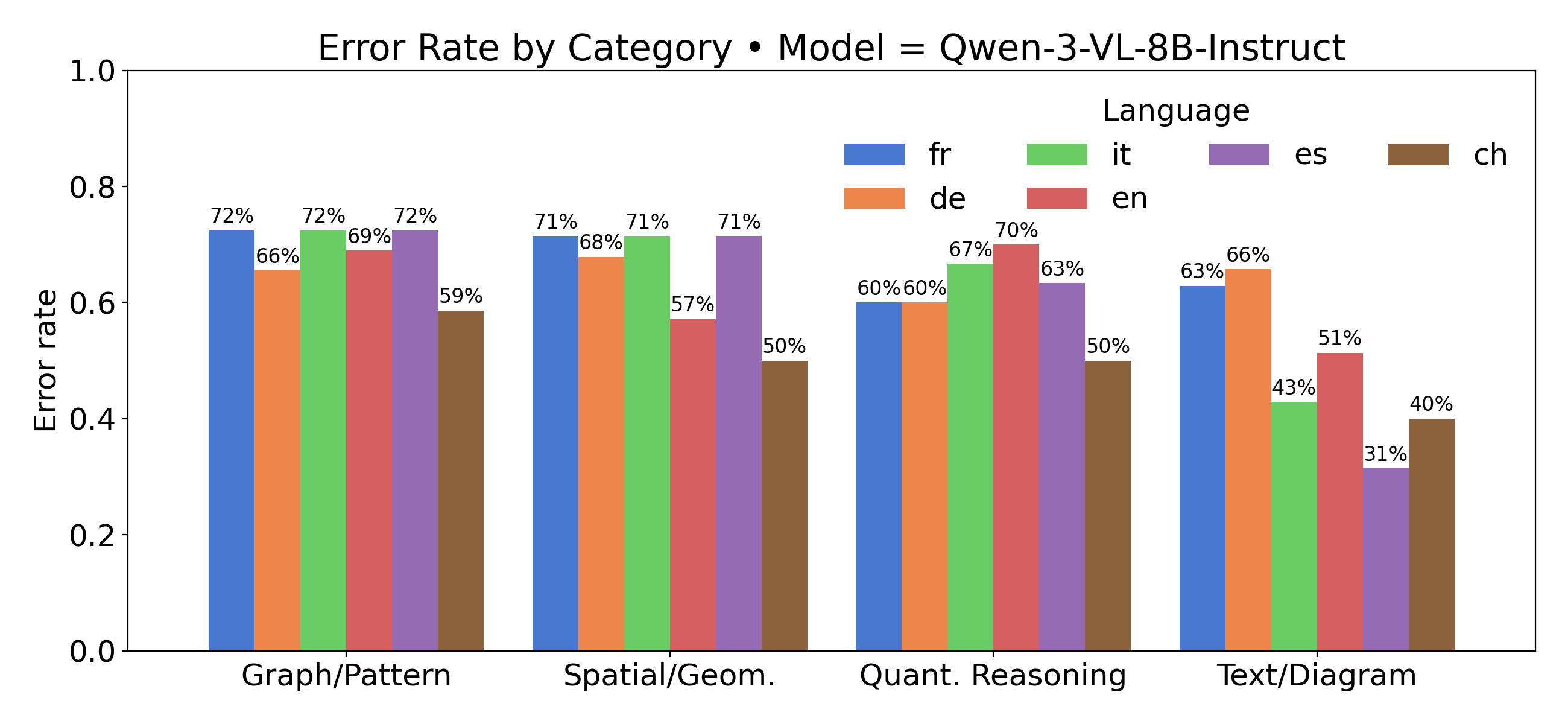}
    \caption{Qwen-3-VL-8B-Instruct}
    \label{fig:error_qwen}
  \end{subfigure}
  \hfill
  \begin{subfigure}[b]{0.48\textwidth}
    \centering
    \includegraphics[width=\textwidth]{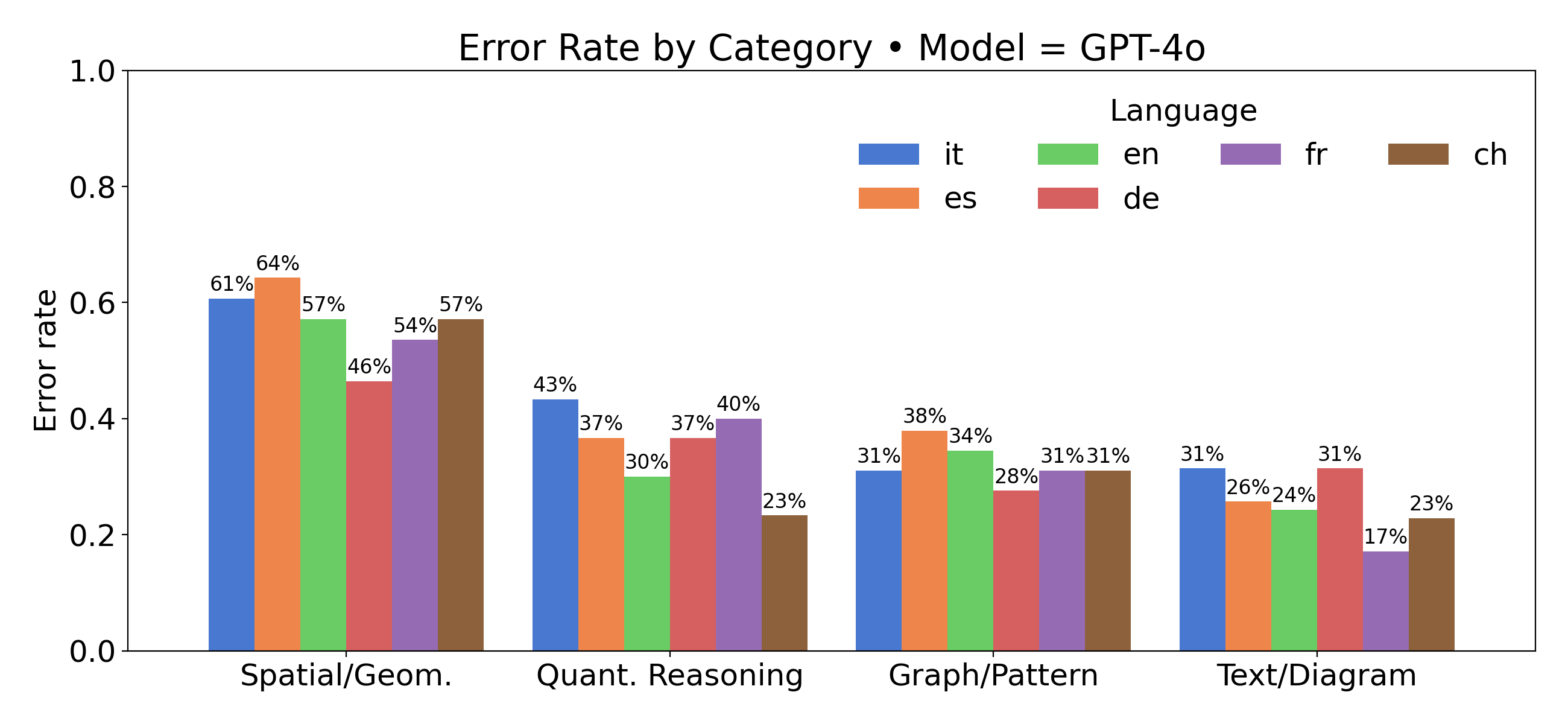}
    \caption{GPT-4o}
    \label{fig:error_gpt4o}
  \end{subfigure}
  \caption{Comparison of error rates across languages for GPT-4o and Qwen-3-VL-8B-Instruct.}
  \label{fig:error_all}
\end{figure*}

\paragraph{Results on Official PISA Scale.}
The official metric used in the international PISA assessments is not the average solve rate but a standardized scale, typically ranging from 350 to 650 points, which enables comparisons across countries. While comparing vision–language models directly with the performance of 15-year-old test takers on this scale would provide valuable insights, the parameters required for the underlying Rasch model are not widely available. However, we were able to obtain item parameters for a subset of questions included in \datasetname{} and report the corresponding results in \Cref{sec:pisa_scores_appendix}.
We find that vision–language models perform worse than human test takers in the mathematics category. Nevertheless, these findings should be interpreted with caution, as the available item difficulty parameters cover only a subset of questions. Further, the official PISA parameters are estimated based on the complete set of test items, whereas our benchmark consists only of a subset of those.

\paragraph{Comparison to Related Benchmarks.}

In~\Cref{tab:bench-compare}, we compare \datasetname{} with commonly known benchmarks MMLU \citep{hendrycks2021measuringmassivemultitasklanguage} and MMMU \citep{yue2024mmmu}. We observe that our benchmark is significantly more difficult compared to MMLU for the models investigated, e.g., we find a performance of 68.5 on MMLU using Qwen2.5-VL (7B), whereas the corresponding performance on \datasetname{} is only 52.4. However, we note that MMLU is a text-only benchmark, emphasizing that modalities other than text is more difficult for current language models.

When considering the results for MMMU, a corresponding multimodal benchmark, we observe that the results on our benchmark are in a similar range. For instance, we find that Qwen2.5-VL (72B) achieves 70.2 on MMMU and 69.4 on \datasetname{}, and similarly Qwen3-VL (30B-A3B) achieves 59.0 on MMMU and 57.3 on \datasetname{}.

\subsection{Error Analysis}
As described in~\Cref{sec:modular_info_extraction}, we labeled each question with one of the following question type categories: spatial and geometric reasoning, quantitative reasoning, graph and pattern analysis, and text and diagram understanding. In this section, we investigate the errors the models have made to better understand the failure areas.

We show results for Qwen2.5-VL (7B) in~\Cref{fig:error_qwen} and for GPT-4o in~\Cref{fig:error_gpt4o}. For Qwen-3-VL (8B), we observe high error rates, especially in the graph \& pattern analysis categories, of up to 72\% for French, Italian, and Spanish. The error rates in quantitative reasoning and spatial \& geometric reasoning are slightly lower on average compared to graph \& pattern analysis. However, the lowest error rate in these categories is still at 50\% for Chinese. The category with the lowest error rates on average is text and diagram understanding.

These trends also hold when looking at the results for GPT-4o; however, all error rates are significantly lower on average, except for the spatial \& geometric reasoning category. Particularly for the categories quantitative reasoning and graph \& pattern analysis, we observe substantially lower error rates. For example, the error rate in quantitative reasoning in Italian drops from 67\% to 30\%, or the error rate in graph \& pattern analysis drops from 72\% to 31\% in French.

\subsection{Contamination Analysis}
In this section, we investigate to what extent models may have memorized parts of the benchmark during pretraining, as our data source is a publicly available resource. To do so, we conduct a contamination analysis using two settings in which we evaluate the model with and without the images.
Our hypothesis here is that if a model can correctly answer questions without the visual context, it indicates prior exposure to the same or similar PISA tests during pretraining, which would undermine the validity of the benchmark. We conduct the experiment using the original English split and all translated versions of our benchmark.

\begin{table*}[ht]
    \centering
    \begin{tabular}{lcccccc|c}
    \toprule
    & \multicolumn{7}{c}{\textsc{Accuracy}} \\
    \textsc{Model} & \textsc{ch} & \textsc{de} & \textsc{en} & \textsc{es} & \textsc{fr} & \textsc{it} & \textsc{Avg.} \\
    \midrule
    \textit{With Images} \\
    Qwen2-VL-7B-Instruct & 49.56 & 57.52 & 53.98 & 64.60 & 61.95 & 52.21 & 56.64 \\
    Qwen2.5-VL-7B-Instruct & 53.10 & 55.75 & 61.06 & 64.60 & 60.18 & 65.49 & 60.03 \\
    \midrule
    \textit{Without Images} \\
    Qwen2-VL-7B-Instruct & 32.74 & 26.55 & 35.40 & 40.71 & 44.25 & 31.86 & 35.25 \\
    Qwen2.5-VL-7B-Instruct & 36.28 & 29.20 & 33.63 & 34.51 & 38.94 & 36.28 & 34.81 \\
    \bottomrule
    \end{tabular}
    \caption{Contamination test results comparing model performance with and without access to images.}
    \label{tab:contamination_results}
\end{table*}

We depict the results in~\Cref{tab:contamination_results}. The results show a substantial decrease in accuracy across all languages and models when images are removed, indicating our benchmark does not suffer from pretraining contamination.
For example, for Qwen2.5-VL-7B-Instruct, average accuracy decreases from 60.0\% to 34.8\%, and for Qwen2-VL-7B-Instruct, from 56.6\% to 35.3\%.  

This consistent decline demonstrates that models depend heavily on the visual input to answer correctly. If prior exposure to our dataset were present, models would be expected to achieve accuracy levels comparable to the text-only setting by recalling question–answer pairs from pretraining. Instead, the large performance gap between the two settings confirms that models have not been trained on the benchmark content and that their predictions require visual reasoning.

These findings, together with the consistently low to moderate overall scores, indicate that \datasetname{} exhibits low contamination and provides a reliable measure of multimodal reasoning. A likely explanation for the low contamination level is that each input question was reframed and partially adjusted to fit a multiple-choice format, making it more difficult for models to reproduce memorized answers from pre-training data.

\section{Related Work}

\paragraph{LVLM Benchmarks.}
Recent progress in large vision-language models (LVLMs) has been closely tied to the development of evaluation benchmarks. Early benchmarks primarily assessed visual perception and image understanding \cite{fu2023mme,MMBench,liu2023visualinstructiontuning,meng2024mmiumultimodalmultiimageunderstanding}, often restricting evaluation to multiple-choice or short-form VQA tasks. More recently, these benchmarks have been extended to more general areas such as cognition and reasoning, for example, benchmarks such as M3Exam \citep{zhang2023m3exammultilingualmultimodalmultilevel}, SOK-Bench \citep{wang2024sokbenchsituatedvideoreasoning}, MathVista \citep{lu2024mathvistaevaluatingmathematicalreasoning}, or VL-ICL-Bench \citep{zong2025vliclbenchdevildetails}. Further, benchmarks like MMDU \cite{liu2024mmdu}, adopt open-ended questions combined with LLM-as-a-judge evaluation, while MMMU-pro \cite{yue2024mmmu} propose unifying text and images into a single visual representation.

\paragraph{Multilingual Benchmarks.}
In the domain of multilingual benchmarks, authors often begin with English datasets and translate them into other languages, such as XNLI \citep{conneau-etal-2018-xnli} or XCOPA \citep{ponti-etal-2020-xcopa}. More recent works utilize multilingual language models to translate the original test set, such as HumanEval-XL \citep{peng2024humanevalxlmultilingualcodegeneration} or mHumanEval \citep{raihan-etal-2025-mhumaneval}. However, while this approach provides broad coverage, it may propagate cultural or linguistic biases through the translation using language models \cite{shi2022language}. P-MMEval \citep{zhang2024p} and BenchMAX \citep{huang2025benchmax} address this issue by using parallel corpora to fairly assess cross-lingual capabilities, disentangling cultural knowledge from a language model's translation ability.

\paragraph{Multilingual LVLM Benchmarks.}
Many multilingual LVLM benchmarks evaluate the general natural language and image understanding capabilities of vision-language models such as xGQA \citep{pfeiffer2021xgqa}, GEM \citep{su2021gemgeneralevaluationbenchmark}, and MaXM \citep{changpinyo2022maxm}. More recent benchmarks extend this line of work to reasoning and cognition, including M3Exam \citep{zhang2023m3exammultilingualmultimodalmultilevel}, EXAM-V \citep{das2024exams}, or M5-VGR \citep{schneider2024m5}. Others focus on culture-specific reasoning, such as ALM-bench \citep{vayani2024all} and CVQA \citep{romero2024cvqa}, or domain-specific reasoning, such as medical reasoning in WorldMedQA-V\citep{matos2024worldmedqavmultilingualmultimodalmedical}. To mitigate the challenges of translating datasets into target languages, parallel benchmarks such as M4U \citep{wang2024m4u}, PM4Bench \citep{gao2025pm4benchparallelmultilingualmultimodal}, and XT-VQA \citep{yu2024cross} have been proposed.

\section{Conclusion}
In this paper, we introduce \datasetname{}, a multilingual and multimodal benchmark designed to evaluate vision-language models on human-authored tests based on the international PISA study of the OECD. We translated the original English set using GPT-4 and verified the translation accuracy with native speakers. Further, we enrich our dataset with question type categories that enable the detailed analysis of failure areas. 

In our experiments, we find that state-of-the-art vision-language models struggle to achieve high accuracy rates across all languages. We further observe significant gaps when evaluating on our non-English splits, highlighting the need for better approaches to multilingual and multimodal reasoning. At last, we find particularly high error rates in the category of geometric and spatial reasoning, indicating that this area is still challenging for current state-of-the-art VLMs. 

%By publicly releasing \datasetname{} and its evaluation framework, we provide the research community with a robust and reproducible benchmark for multilingual multimodal reasoning based on the international PISA tests.

%By making \datasetname{} publicly available, we provide the community with a resource to rigorously evaluate progress in multilingual multimodal reasoning. We believe this dataset will encourage the development of models that are more robust, equitable, and better aligned with real-world tasks. In future work, we plan to extend \datasetname{} to additional languages and domains, further broadening its scope as a benchmark for comprehensive multimodal evaluation.

% \clearpage

\section*{Limitations}

% GPT contamination as a limitation

While \datasetname{} provides a valuable resource for evaluating multilingual multimodal reasoning, it also comes with several limitations. First, we do not perform the translations ourselves; instead, we rely on human annotators to verify and correct automatically generated translations. Although this process ensures sufficient quality for evaluation, it may not capture all subtle linguistic nuances, particularly in languages with complex morphology or idiomatic expressions. 

Second, our benchmark uses accuracy rates as our main evaluation metric. However, the underlying PISA tests are designed to compare 15-year-old students across countries. As a result, the official test metrics are PISA scores, calculated using item response theory. We were only able to get the difficulty scores for a subset of our questions, such that we could only estimate the PISA scores in \Cref{sec:pisa_scores_appendix}.

Third, although the dataset consists of high-quality, human-authored examples, its size remains relatively modest. This makes \datasetname{} particularly suitable as a resource-friendly evaluation benchmark, but less suitable for extensive and fine-grained error categorization or large-scale model training. Future extensions could address this limitation by expanding the dataset with additional PISA examples or complementary educational resources.

Fourth, during dataset creation, we observe that translating the images leads to substantial performance decreases, largely due to the inability of multilingual VLMs to accurately translate the visual content. We identify several major errors, including incorrect unit conversions, hallucinated image descriptions, and omissions of essential information. In future work, we plan to source the images directly from the PISA tests in each respective language to ensure accurate, human-performed translations.

Finally, our evaluation protocol currently relies on LLM-as-judge evaluation. While this evaluation approach exploits the generation capabilities of LLMs, they may still miss subtler reasoning errors or reward overgeneralized answers. More robust evaluation methods, such as rubric-based scoring or task-specific human assessment, could further strengthen conclusions drawn from \datasetname{}.

% Access to PISA tests for other languages than German, Spanish, and English. Perhaps, with the publication of this Paper, the OECD will make more PISA PDFs available for research purposes. 

% A Small group of human annotators leading to human bias in the translation verification. 

% Machen wir nicht! 
% \section*{Ethics Statement}

\section*{Acknowledgments}
This paper has received funding from the European Union’s Horizon Europe research and innovation programme under grant agreement No. 101189745 (HIVEMIND) and the German Research Foundation (DFG) project NFDI4DS under Grant No.: 460234259. We would also like to thank Yuan Li and her team from the Institute of German Studies at Zhejiang University, as well as Martin Courtois, Carlos Franzreb, and Alessandro Noveta, for annotating the dataset and verifying the translations.

Alan Akbik and Patrick Haller are supported by the Deutsche Forschungsgemeinschaft (DFG, German Research Foundation) under Emmy Noether grant ``Eidetic Representations of Natural Language'' (project number 448414230). Further, Alan Akbik is supported under Germany’s Excellence Strategy ``Science of Intelligence'' (EXC 2002/1, project number 390523135). Jonas Golde is supported by the Bundesministerium für Bildung und Forschung (BMBF) as part of the project ``FewTuRe'' (project number 01IS24020).

%We thank all reviewers for their valuable comments. Jonas Golde is supported by the Bundesministerium für Bildung und Forschung (BMBF) as part of the project ``FewTuRe'' (project number 01IS24020). Alan Akbik and Patrick Haller are supported by the Deutsche Forschungsgemeinschaft (DFG, German Research Foundation) under Emmy Noether grant ``Eidetic Representations of Natural Language'' (project number 448414230). Further, Alan Akbik and Max Ploner are supported under Germany’s Excellence Strategy ``Science of Intelligence'' (EXC 2002/1, project number 390523135).

%\nocite{*}
\section*{Bibliographical References}
\label{sec:reference}

\bibliographystyle{lrec2026-natbib}
\bibliography{references}

\newpage
\appendix

\section{Extraction Prompt} \label{sec:extraction_prompt_appendix}

\Cref{fig:extraction_prompt} shows the prompt used for our information extraction step, especially for the data augmentation part, to ensure consistency and self-containment of each question.

\begin{figure*}[h]
\centering
\begin{tcolorbox}[colback=gray!5!white,
                  colframe=black!75!black,
                  title={Prompt Template for Question Extraction from PDFs},
                  left=2mm,right=2mm,top=1mm,bottom=1mm,
                  width=\textwidth]
                 
You are an expert educational content creator and a skilled test designer. Your task is to analyze the provided PDF document, which contains educational content, exercises, and questions.
\
\
Your goal is to perform the following actions for every question or exercise you find in the document:
\
\
1.  \textbf{Extract the Core Questions}: Identify the primary questions or problem statements.\\
2.  \textbf{Standalone Questions}: Ensure each question is self-contained, providing all necessary context for understanding without needing to refer back to the document. Assume only tables and images are provided separately if referenced.\\
3.  \textbf{Reformulate into a Multiple-Choice Question (MCQ)}: Convert the question into a standard multiple-choice format with exactly four options (A, B, C, D).\\
4.  \textbf{Generate a Correct Answer}: Identify the correct solution from the context of the document and assign it to one of the options. This can include a reference image if the question is visual in nature. Assume that the image is available to the test-taker, but refer to it in the question text.\\
5.  \textbf{Generate Plausible Distractors}: Create three plausible but incorrect answer choices (distractors) that are relevant to the topic but are definitively wrong. The distractors should be well-formed and not nonsensical.\\
6.  \textbf{Identify Scoring}: If the original question has a scoring system (e.g., "Worth 10 points"), you must extract and include this information. If no score is present, state "Score: Not specified."\\
7. \textbf{Translate if needed}: If the document is not in English, translate the question and options into English while preserving the original meaning. Apply also to the task name if it is not in English.\\
8. \textbf{Category Identification}: Identify whether the question falls under "math" or "reading" and specify a more detailed subcategory if possible (e.g., geometry, algebra for math; locate info, interpret text for reading).\\
\
\
The final output must be a JSON array, where each object represents one multiple-choice question. A PDF can contain more than one question. If there is more than one question, then identify each and produce one JSON object per question. 
\end{tcolorbox}
\caption{Prompt used for augmenting the original English test questions, e.g., required when multiple-choice answer options are not available or the question is not self-contained.}
\label{fig:extraction_prompt}
\end{figure*}

\section{Translation Prompt} \label{sec:translation_prompt_appendix}

We translate all English examples of our dataset using GPT-4 into our five target languages: Spanish, German, Chinese, French, and Italian. We use OpenAI's platform to generate the translation and depict the prompt used in~\Cref{fig:translation_prompt}.

\begin{figure*}[h]
\centering
\begin{tcolorbox}[colback=gray!5!white,
                  colframe=black!75!black,
                  title={Prompt Template for Translations},
                  left=2mm,right=2mm,top=1mm,bottom=1mm,
                  width=\textwidth]
          You are an expert translator from English to \{lang\} for educational assessments. 
You are given a multiple-choice question (MCQ) in English with its answer %options and the correct answer.\\
Translate the question and all answer options into <lang>, BUT keep all units, symbols, labels, and tokens exactly as in the original (no localization), so the text aligns with English annotations present in the associated image.\\

Requirements:\\
1. \textbf{Literal alignment with the source}:\\
   - Do NOT convert or localize measurement units or quantities (e.g., keep 'miles per hour', 'mph', 'inches', 'pounds', 'Fahrenheit' exactly as written).\\
   - Preserve all numbers, formulas, variables, dates, and proper nouns exactly (do not change numeric formatting: keep 3.5, not 3,5).\\
   - Preserve abbreviations, acronyms, and labels verbatim if they may appear in the image (e.g., 'mph', 'NYC', axis labels, map keys).\\
\
2. \textbf{Natural but faithful \{lang\}}:\\
   - Translate the surrounding prose naturally into {lang} but do NOT alter difficulty or meaning.\\
   - Do NOT add hints, explanations, or paraphrases that change the task.\\
\\
3. \textbf{Preserve the answer key}:\\
   - Ensure the translated correct option remains the correct answer in \{lang\}.\\
   - Do NOT reorder options or change their content beyond the literal translation.\\

4. \textbf{Output format}: Return ONLY the translated JSON in this exact structure, as you received it.\\
\
Do not add any extra fields, comments, or explanations. If a term cannot be translated without breaking alignment with the image, keep it in English verbatim.\\
\end{tcolorbox}
\caption{Prompt template used for generating translations using GPT-4.}
\label{fig:translation_prompt}
\end{figure*}

\section{Approximating Model PISA Scores} \label{sec:pisa_scores_appendix}

PISA test scores are not reported as solve ratios but are instead measured on the PISA scale using item response theory (IRT) \citep{hambleton2013item,okubo2022theoretical}. To enable a direct comparison between model performance and student ability on the PISA scale, we estimate an approximate PISA-equivalent score using a Rasch model formulation \citep{rasch1980probabilistic}. We were able to find the official PISA difficulty parameters for a subset of questions in \datasetname{}. We treat the model's binary correctness outcomes as responses to Rasch-parameterized items and infer the latent ability parameter $\theta$ by maximizing the Bernoulli log-likelihood under the logistic Rasch model. This inferred ability is then mapped to the PISA scale, providing an interpretable, though imperfect, score, as such scores are not widely publicly available. However, it helps us to establish a link between model accuracy on real assessment items and the corresponding human performance level. 

We present the PISA of vision-language models in \Cref{tab:pisa_points}. Since only a limited number of benchmark items include published PISA item parameters, the resulting estimates rely on a comparatively small sample. As such, these PISA-equivalent scores should be interpreted as \textit{indicative trend signals} rather than firm psychometric measurements. To ensure numerical stability, we constrain the Rasch ability parameter to $[-3, 3]$, which corresponds to roughly the 200-800 point range of the PISA scale. Under these bounds, a model answering every available item correctly attains the upper-limit score of 800, which is achieved by GPT-4o in the reading area. While this provides a useful approximation for positioning models along the PISA scale, a more directly comparable metric would be the per-item student accuracy distributions reported by PISA (i.e., the proportion of students solving each question correctly), which, unfortunately, are not publicly available for all released items. 
We therefore include the Rasch-based scores for transparency and illustrative comparison with human performance, while emphasizing that they should not be interpreted as evidence that language models currently match or exceed student proficiency on the full PISA assessment. Due to limited PISA scores assigned to our corresponding samples, we aggregated over years.

 To enable a direct comparison between model performance and student ability on the PISA scale, we estimate an equivalent PISA score using a Rasch model formulation. For the subset of benchmark items where official PISA difficulty estimates were available, we treat the model’s binary correctness outcomes as responses to Rasch-parameterized items and compute the latent ability parameter $\theta$.

\begin{table*}[b]

\resizebox{\textwidth}{!}{
\begin{tabular}{lccccccccccccrr}
\toprule
\textsc{Language} & \multicolumn{2}{c}{\textsc{en} \flagUSA} & \multicolumn{2}{c}{\textsc{de} \flagGER} & \multicolumn{2}{c}{\textsc{fr} \flagFRA} & \multicolumn{2}{c}{\textsc{it} \flagITA} & \multicolumn{2}{c}{\textsc{es} \flagESP} & \multicolumn{2}{c}{\textsc{ch} \flagCHN} & \multicolumn{2}{c}{Avg.} \\
\textsc{Category} & \textsc{Math} & \textsc{Reading} & \textsc{Math} & \textsc{Reading} & \textsc{Math} & \textsc{Reading} & \textsc{Math} & \textsc{Reading} & \textsc{Math} & \textsc{Reading} & \textsc{Math} & \textsc{Reading} & \textsc{Math} & \textsc{Reading} \\
\textsc{Model} &  &  &  &  &  &  &  &  &  &  &  &  & & \\
\midrule
GPT-4o & 576 & 800 & 559 & 800 & 542 & 800 & 559 & 800 & 542 & 800 & 576 & 800 & 559 & 800 \\
Claude-3-5-Haiku & 632 & 800 & 559 & 650 & 677 & 604 & 593 & 604 & 559 & 604 & 576 & 604 & 599 & 644 \\
\midrule
Qwen2.5-VL-3B-Instruct & 444 & 656 & 444 & 650 & 494 & 604 & 426 & 604 & 461 & 604 & 426 & 604 & 449 & 620 \\
Qwen2.5-VL-7B-Instruct & 526 & 718 & 526 & 566 & 542 & 800 & 526 & 800 & 542 & 650 & 494 & 566 & 526 & 683 \\
Qwen2.5-VL-32B-Instruct & 526 & 610 & 444 & 650 & 461 & 713 & 408 & 800 & 444 & 604 & 444 & 650 & 454 & 671 \\
Qwen2.5-VL-72B-Instruct & 593 & 718 & 559 & 800 & 542 & 800 & 612 & 800 & 542 & 800 & 576 & 800 & 570 & 786 \\
\midrule
Qwen3-VL-4B-Instruct & 526 & 656 & 494 & 650 & 559 & 604 & 542 & 713 & 478 & 650 & 510 & 650 & 518 & 653 \\
Qwen3-VL-8B-Instruct & 526 & 800 & 526 & 650 & 542 & 650 & 494 & 713 & 494 & 800 & 526 & 650 & 518 & 710 \\
Qwen3-VL-30B-A3B-Instruct & 576 & 718 & 461 & 604 & 494 & 713 & 444 & 800 & 426 & 532 & 494 & 713 & 482  & 680 \\
\midrule
gemma-3-4b-it & 461 & 456 & 426 & 501 & 461 & 470 & 461 & 470 & 444 & 532 & 478 & 470 & 455 & 483 \\
gemma-3-12b-it & 526 & 800 & 542 & 800 & 526 & 713 & 510 & 650 & 526 & 650 & 526 & 800 & 526 & 735 \\
gemma-3-27b-it & 576 & 656 & 542 & 650 & 576 & 650 & 576 & 604 & 542 & 650 & 494 & 566 & 551 &  629 \\
\midrule
llava-1.5-7b-hf & 366 & 429 & 387 & 470 & 341 & 604 & 408 & 650 & 426 & 532 & 426 & 501 & 392 & 531 \\
llava-1.5-13b-hf & 408 & 512 & 408 & 470 & 366 & 470 & 341 & 532 & 366 & 470 & 387 & 501 & 379 & 492 \\
llava-v1.6-34b-hf & 408 & 718 & 408 & 650 & 408 & 650 & 387 & 713 & 408 & 713 & 461 & 650 & 413 & 682 \\
\midrule
SmolVLM2-256M-Video-Instruct & 387 & 429 & 494 & 373 & 510 & 373 & 510 & 470 & 526 & 440 & 461 & 373 & 481 & 409 \\
SmolVLM2-500M-Video-Instruct & 313 & 337 & 387 & 408 & 341 & 281 & 387 & 332 & 408 & 373 & 341 & 440 & 362 & 361 \\
SmolVLM2-2.2B-Instruct & 461 & 456 & 387 & 332 & 366 & 373 & 387 & 408 & 444 & 440 & 387 & 440 & 405 & 408 \\
\midrule
Idefics3-8B-Llama3 & 494 & 610 & 444 & 650 & 478 & 470 & 461 & 532 & 461 & 566 & 387 & 501 & 454 & 554 \\
\midrule
%\footnote{Scores from 2022: https://worldpopulationreview.com/country-rankings/pisa-scores-by-country (accessed 24.10.2025)}
\textit{Test Takers} \\
2000 & 493 & 504 & 490 & 484 & 517 & 505 & 457 & 487 & 476 & 493 & - & - & 487 & 495 \\
2003 & 483 & 495 & 503 & 491 & 511 & 496 & 466 & 476 & 485 & 481 & - & - & 490 & 488 \\
2006 & 474 & - & 504 & 495 & 496 & 488 & 462 & 469 & 480 & 461 & - & - & 483 & 478 \\
2009 & 487 & 500 & 513 & 498 & 497 & 497 & 483 & 486 & 483 & 481 & 600 & 556 & 511 & 503 \\
2012 & 481 & 498 & 514 & 508 & 495 & 505 & 485 & 490 & 484 & 488 & 613 & 570 & 512 & 510 \\
2022 & 465 & 504 & 475 & 480 & 474 & 474 & 471 & 482 & 473 & 474 & 552 & 510 & 485 & 487 \\
% Human\footnote{Scores from 2022: https://worldpopulationreview.com/country-rankings/pisa-scores-by-country (accessed 24.10.2025)} & \emoji{flag-united-states} 465 & \emoji{flag-united-states} 504 & \emoji{flag-germany} 475 & \emoji{flag-germany} 480 & \emoji{flag-france} 474 & \emoji{flag-france} 474 & \emoji{flag-italy} 471 & \emoji{flag-italy} 482 & \emoji{flag-spain} 473 & \emoji{flag-spain} 474 & \emoji{flag-china} 552 & \emoji{flag-china} 510 & 485 & 487 \\
\bottomrule
\end{tabular}}
\caption{Approximate PISA-equivalent model scores compared with national averages for the United States, Germany, Italy, Spain, France, and China from the past years. We compute the PISA scores for LLMs on a subset of questions for which we were able to obtain difficulty parameters. Thus, the resulting LLM scores should be interpreted as approximate indicators rather than validated test scores.}
\label{tab:pisa_points}
\end{table*}

\end{document}